# Soft Computing Techniques for Change Detection in remotely sensed images : A Review

Ms Madhu Khurana[1] and Dr Vikas Saxena[2]

[1]Dept. Of Computer Science & Engineering, ABES Engineering College
Ghaziabad-201009, India

[2]Department of CS & IT, Jaypee Institute of Information Technology,
Noida – 201307, India

## Abstract

With the advent of remote sensing satellites, a huge repository of remotely sensed images is available. Change detection in remotely sensed images has been an active research area as it helps us understand the transitions that are taking place on the Earth's surface . This paper discusses the methods and their classifications proposed by various researchers for change detection. Since use of soft computing based techniques are now very popular among research community, this paper also presents a classification based on learning techniques used in soft-computing methods for change detection.

## 1. Introduction

In Remote Sensing, change detection means assessing or measuring the change on the Earth's surface by jointly processing multi-temporal images of the same geographical area acquired at different times. This field has attracted a lot of effort in research due to its applications in various areas as Land-use and land-cover (LULC), Forest or Vegetation change, Forest mortality, defoliation and damage assessment, Wetland change, Urban expansion, Damage assessment, Crop monitoring, Changes in glacier mass balance, Environmental change and Deforestation, regeneration & selective logging[1]. Change detection (CD) on earth's surface is an active research topic since it can help in monitoring and optimal planning of Earth's resources and also help to arrest undesired changes. Any change detection system should be able to (a) define the change area and change rate; (b) distribution of change areas; (c) change trajectories; and (d) the accuracy assessment of the change detection methods. [2]

The result of any CD is affected by various factors s.a. spatial, spectral, thematic and temporal constraints; radiometric resolution, atmospheric conditions and soil moisture conditions. The methods being developed are usually application specific. Various techniques have been developed for CD using both pixel based as well as object based approaches.

## 2. Change Detection Process : An overview

The change detection systems designed use the data generated by different satellites having different radiometric resolutions, spatial resolution & orbiting frequency. Most commonly SPOT (Systeme Pour l'Observation de la Terre(SPOT) satellite system), NOAA's AVHRR (Advanced Very High Resolution Radiometer, http://www.noaa.gov/), MODIS (Moderate Resolution Imaging Spectroradiometer, http://modis.gsfc.nasa.gov/index.php), LANDSAT (http://landsat.gsfc.nasa.gov/?page_id=9) data is used. Also very high resolution data from optical satellites s.a QuickBird, IKONOS, EROS and GeoEye can also be used. For CD systems to give good results, the multi-temporal images taken should be of the same spatial and radiometric resolution, and proper registration of the images should be done, so as to avoid false alarms. Also the images should have near anniversary acquisition dates, so as to overcome the problems of sun angle and phenological differences.

As discussed by D. Lu et. al in [2], the process of change detection includes three major steps

Step 1: **Image preprocessing** –Before applying any Change Detection algorithm on the multitemporal images, we need to carry radiometric corrections and image registration, so as to increase the accuracy of the CD. It may also include geometric rectification, atmospheric corrections and also topographic corrections, if the terrain contains mountains. Radiometric correction is done to reduce the inconsistency between the values surveyed by sensors arising due to different imaging seasons, different solar altitudes, different angles or different cloud or snow covers[11]. Also image registration of multi-temporal images is required so as to avoid false change detection for pixel-based CD methods. Image registration may not be necessary for feature-based CD methods where extracted features can be compared for change detection[11]. Jianya (2008) suggested that image registration and change detection should be integrated in a single algorithm.

Step 2 : **Change Detection Algorithm or technique** : we choose a technique or algorithm to detect and analyze the change. Selecting an algorithm for change detection is a difficult task. Different algorithms can be studied and the algorithm best suited for the application area, can be selected, depending on the kind of results expected. If the application only requires whether a change has occurred or not, then some set of algorithms may be suited while if we also want to work on change area and the direction of change, another set may be suitable. The algorithm will also depend upon the radiometric and spatial resolutions of the images, some pixel based method may not be suited for VHR images[4]. The spatial resolution dictates the kind of study that can be performed on the RS data, if the resolution is coarse the change over a large area only can be done, if we wish to perform a sub-metric change detection, then the data available should support such resolutions.

Step 3 : **Accuracy Assessment** : Accuracy assessment determines the quality of information retrieved from remotely sensed data. We can have a qualitative or a quantitative accuracy assessment. In qualitative assessment we assess the accuracy by looking at the map and comparing it with what we see on the ground while in case of quantitative assessment we try to identify and measure the remote sensing error by comparing it with the ground truth.[11]

Designing a change detection system requires a deliberate effort to understand the kind of changes that need to be observed. Deciding on the period for which data is required (or temporal difference between the data images), spectral resolution and the nature of result expected will guide in data collection. If the changes to be observed cover a large geographical area (like land cover – land change), the images with less radiometric resolution will suffice, else in case we need to observe the changes in detail (sub-metric), high resolution images would be required. Images from different resolutions can also be combined in an application, using GIS software, here pre-processing would be required to bring the images at same resolution, followed by change detection.

## 3. Existing Classifications of CD Techniques

In the past different classifications of the CD methods have been proposed. This section discusses some proposed classifications. A. Singh [3] in his paper on review of the digital change detection techniques in 1989 categorized the change detection research on the basis of (a) The data transformation used & (b) the analysis technique used to detect the change. He classified the research work done till that time into various categories. A brief overview of all the prevalent techniques s.a. image differencing, image rationing, image regression, vegetation index differencing , principal component analysis, post-classification comparison and Direct multidate classification, Change Vector Analysis (CVA) was offered. While evaluating these techniques on the basis of literature reviewed, it was concluded that different CD methods produced different change maps. It was observed that most of the results were not compared with the ground truth and thus the capability of these methods were poorly evaluated. The paper provided the best results for each technique of CD. The accuracy for these techniques was in the range of 51% to 74%.

Jianya et. al[11] in 2008 proposed two broad groups for classifying the CD methods, namely, bi-temporal and temporal trajectory analysis. Bi-temporal methods measures the change between two date images while temporal trajectory analysis analyses the change based on continuous timescale measuring not only the change between two dates but also the progress of change over the period. Further he gave seven categories for CD methods, namely, direct comparison, classification, object-oriented method, model method, time-series analysis, visual analysis and hybrid method.

M.Hussain et. al[4] in a recent paper(2013) have reviewed and categorized the change detection methods based on the unit of analysis, i.e. pixel based and object based approaches. In the pixel based methods the analysis is done on the basis of variation in pixel intensities while object based approaches, first extract objects from the image and then try to perform change detection based upon the objects extracted. The object based methods are more suitable for Very High Resolution (VHR) images. The subclasses with each category are defined and the CD methods are categorized based upon these classes. Figure 1 is the pictorial representation of the classification provided by the authors.

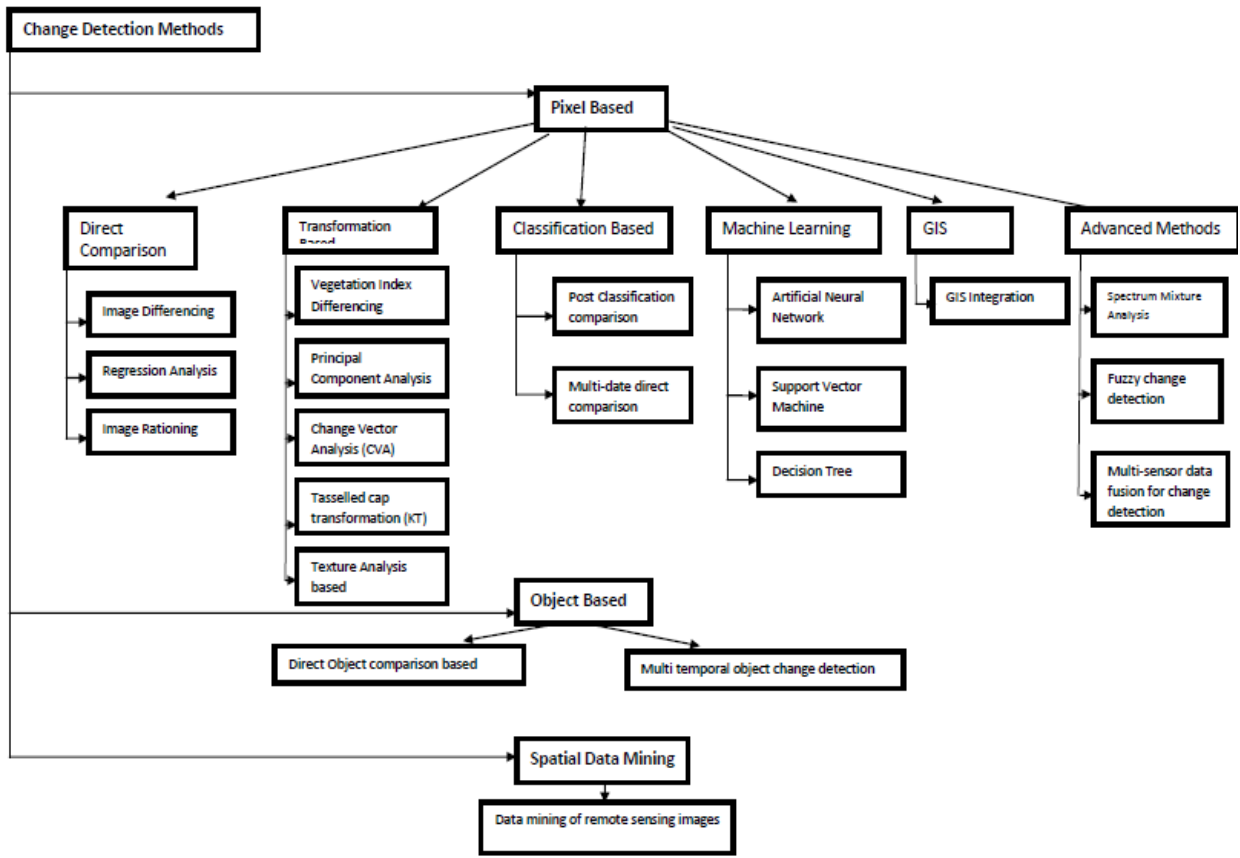

Figure 1 : Classification of change detection methods(As in M.Hussain et. al (2013))

The advantages and disadvantages as well as examples where these method have been used, has been discussed by Hussain et.al(2013), the interested readers can refer it for more details.

## 4. Soft Computing Based Change Detection Methods

Soft computing is a fusion of methodologies designed to model and enable solutions to real life problems, which are not modeled or too difficult to model mathematically. It has the capability to tolerate imprecision, uncertainty, approximate reasoning and partial truth. Soft computing is a collection of Evolutionary computing, neural networks(NN) (including neural computing), fuzzy systems (including fuzzy logic(FL)) , machine learning(ML), and probabilistic computing(PC). Evolutionary computing(EC) encompasses Genetic Algorithms, Evolutionary strategies & Evolutionary programming.

Change detection methods using soft computing can be divide into various categories depending upon learning paradigm being used by the system, another distinction can be made on the techniques being used to model to find the solution. Soft computing techniques, NN, EC, ML, FL and PC are complementary to each other rather than competitive. Therefore a combination of these techniques is usually used to model a real life problem. Fuzzy logic plays a critical role as it provides the tolerance for imprecision and allows to take decision in a human like manner.

Change detection(CD) in remote sensing requires large amount of data to be analyzed. Automatic and unsupervised CD techniques can reduce the time required for manual image analysis. Also it is difficult to gather the precise ground truth about the entire regions being covered by the RS images.
The following section provides a classification of CD methods based on soft-computing techniques being used.

# 5 Classification based on Learning Paradigm

5.1 Unsupervised Change Detection

Following pre-processing and calculation of the difference image, we need to classify the data in the difference image into change and no-change sets. Unsupervised CD techniques do not require training data. So the task is to discriminate the data into changed and unchanged sets. The problem reduces to clustering the data into two groups, changed and unchanged. We need to partition the data into disjoint groups where each group contains similar data.

Unsupervised methods can be divided into context-insensitive (based on spectral information) (A.Singh. 1989) and context sensitive(using spatial information)(Ghosh 20014). Histogram equalization is an example of context-insensitive method of classification which does not use neighborhood information(spatial relations). Markov Random Fields are used to overcome the problems of context-insensitive methods. [15]

A number of clustering algorithms can be used[12] :

**Hard C- Means (HCM) or K-means :** The set of patterns are to be divided into c clusters( or k clusters). We randomly chose c number of patterns as the initial cluster centers. Then the distance of each pattern from each cluster center is calculated and we assign the pattern to cluster having minimum distance. Once the assignment has been done for all the patterns in the dataset, we again calculate the new cluster centers by taking the mean of all the patterns in the cluster. Again the distance of each pattern is calculated from the cluster centers and the pattern reassigned to the nearest cluster centers. This process continues till the centers become fixed i.e. there is no change in the cluster centers over an iteration. The distance measure can be defined for a specific problem but usually the Euclidean distance is used. The clusters thus formed can be further analyzed for designating them as change or no-change clusters.

$$O_{kmeans} = \sum_{j=1}^{k} \sum_{X_{mn} \in X_j} \|X_{mn} - v_j\| \qquad (1)$$

**ISODATA (Iterative Self Organizing Data Analysis Technique) :** ISODATA is a method of unsupervised classification. Unlike k-means, no a-priori information about the number of clusters is required. It starts with some random no. of clusters and the clusters can be split or merged depending upon the parameters defined by the user. User gives the parameters to this algorithm as various threshold values. The program iterates until the threshold is reached. Some fast variations of the algorithm are defined, s.a., ISOCLUS by Memarsadeghi N. et. al (2007).

**Fuzzy C-Means (FCM) :** The fuzzy C-Means is a classification technique, where each pattern can have membership in more than one cluster, the degree of membership varies in different clusters. The algorithm uses functional optimization, where it attempts to minimize the objective function defined as :

$$J_m(X; U, V) = \sum_{i=1}^{c} \sum_{k=1}^{n} (\mu_{ik})^m D_{ik} \qquad (2)$$

where m > 1 is the fuzzifier, which controls the fuzziness of the algorithm. The higher the value of m, more fuzzy is the algorithm, mostly the values used are between 1.5 to 2.5. The algorithm outputs clusters with similar shapes, sizes and densities, which might not give good results in various situations.[13] [12][15]

**Gustafson-Kessel clustering (GKC) :** GKC is also a fuzzy clustering technique. It is a modification of FCM, where FCM uses Euclidean distance to measure the distance between the cluster center and the patterns, GKC uses an adaptive distance norm. Gustafson and Kessel suggested an adaptive distance measure using fuzzy covariance matrix. GKC can be used to detect clusters with ellipsoidal cluster, by adapting the cluster shape automatically.[13] [12][15]

5.2 Semi-supervised Change Detection

Unsupervised methods are best suited for change detection applications, since ground truth may not be available in many scenarios. Also labeling the complete training set may not be possible. In case, some data is available, where labels are provided, the advantages of both supervised and unsupervised learning can be utilized. A few labeled training patterns can be used along with unlabeled patterns for learning. Semi supervised learning can improve the classification results, in case of non-availability of the sufficient labeled data [13].

S.Basu[16] proposed that semi-supervised clustering can be improved by defining pairwise constraints i.e. pairs of instances labeled as belonging to same or different clusters. Thus supervision is added in the form of these constraints, which are used to either modify the objective function or the distance measure.

Various semi-supervised algorithms have been explored for the classification of remotely sensed images.

**K-means based algorithms :** As discussed earlier k-means is a very popular algorithm used for clustering the data. A number of modifications have been proposed to the basic k-means algorithm to adapt it for semi-supervised learning, i.e. incorporating some label data to help it learn in a better manner. This section discusses some such variations.

(i) *Seeded k-means :* The initial cluster centers are randomly chosen in standard k-means algorithm, here the cluster centers are initialized by the mean of the seeds belonging to that cluster. The labeled data is only used to find the initial cluster centers. Once the cluster centers have been initialized, the algorithm follows the same steps as standard k-means. The seeds are also reassigned to other clusters, if they are nearest to the cluster, thus restricting the impact of defining seeds only to the initialization step [15].

(ii) *Constrained k-means :* This is same as seeded k-means, with the only difference being that the seeds are not reassigned to any other clusters, i.e., their membership is not re-calculated [15].

(iii) *COP k-means :* This is an extension of standard k-means algorithm, defined by Wagstaff et. al (2001). It defines the labeled information in the form of 'must-link' and 'cannot-link' constraints. Both the constraints have to be satisfied for assigning a pattern to a cluster. The distance of the pattern from all the clusters is determined and arranged in the ascending order based on the distance of the pattern from the cluster. For assigning the pattern to a cluster, the cluster at the top of the list is considered first, if assigning the pattern to this cluster does not violate any constraints, then the pattern is assigned to this cluster otherwise the next cluster on the list is considered. Assigning a pattern to a cluster means that the patterns which were already assigned to this cluster do not have a 'cannot-link' constraint defined for the unlabelled pattern. Also it indicates that the patterns assigned to the different clusters do not have a 'must-link' constraint with the said pattern. In case any of the constraints are violated, the next cluster on the list is checked. The rest of the steps are similar to the standard k-means algorithm, apart from the assignment step.

(iv) *Semi-supervised HMRF k-means :* Basu et. al [16] in 2004 proposed a k-means algorithm based on hidden Markov Random Field (HMRF). HMRF has two components :

a) A hidden field ,L, whose variables are unobservable, i.e., unobservable cluster labels for the patterns
b) An observable set, X, of random variable, i.e. the patterns

The Maximum a-posteriori (MAP) configuration of HMRF model is same as maximizing the posteriori probability $Pr(L|\Psi)$. The hidden random variables $l_i$ are linked with neighborhood $N_i$. The must-link constraints M and cannot-link constraints C define the neighborhood over the hidden labels such that neighbors of a point $x_i$ are all the points which have must-link or cannot-link relations with the said data point. The random field defined over the hidden variables is a Markov Random Field.

The HMRF K-means algorithm, takes as input the set of data points, must-link constraints M , cannot-link constraints C, the distance measure D, and constraint violation costs w and w` . It tries to minimize the objective function. The algorithm iteratively assigns the cluster-labels on the datapoints to minimize the objective function, recalculates the centroids based on assignments, recalculates the distance measure D. The process follows till convergence. The other details of the method can be found in the paper by Basu et. al [16].

(v) *Semi-supervised kernel k-means :* A weighted kernel-based approach for semi-supervised learning was proposed by Kulis et. al. The patterns are mapped to a higher dimensional space with the help of a non-linear kernel and a positive weight is associated with each pattern. The algorithm defined a 'reward' for obeying the constraint instead of a penalty for constraint violation. Whenever the patterns having an associated must-link constraint are assigned the same cluster, the penalty is subtracted from the objective function, thus reducing the cost

**Semi-supervised Support Vector Machine ($S^3VM$) :** Standard Support Vector Machines can handle partially labeled datasets by assuming unlabeled datasets as additional optimization variables. Transductive SVM's (TSVMs) were introduced by Vapnik and Sterin (1977) [17]. TSVM uses both labeled and unlabeled data to discriminate the boundary such that it does not pass through dense regions. Transductive learner is one which works on labeled and unlabeled data and cannot work on unforeseen data. While Vapnik et. al. named it as the transductive, it actually is an inductive learner as the resulting classifier works on the entire data space. Given a set of data, the SVMs classify it such that the hyperplane distinguishing the two classes has the maximum margin. The details of the $S^3$VMs can be found in the paper by Vapnik et. al [17].

5.3 Supervised Change Detection

Supervised change detection methods require the presence of ground truth, i.e. labeling of training set into various so as to use this information for proper classification. Supervised methods are superior in the sense that they can recognize the kind of change. They are robust to the different atmospheric and light conditions at time of acquisition. Many methods discussed by A.Singh and Hussain et.al. in their classifications fall into this category. Post classification methods, direct multi-date classification and kernel based methods fall under this category[15]. A lot of work has been done in CD using these techniques. The interested readers can refer [3],[4],[6] for details of these techniques.

5.3 Active Learning Algorithms

Active learning is to train the classifier on a small set of well-chosen examples. The Active learning algorithms are iterative in nature. At each iteration some labeled samples are provided to the model, the model adapts itself to samples provided and then it presents the user with a set of unlabeled samples and the user ranks them, the model is again adapted with the new set of samples provided. The model keeps on improving its adaptation to the classification problem. Active learning requires interaction between the user and the model. [18] The main focus in active learning is that the ranking of the unlabeled samples be such that it is able to discriminate between the two classes. The algorithms can be divided into three families :

(i) *Committee-based heuristics :* This method quantifies the uncertainty of a pixel by considering a committee of learners[18]. Each member of the committee labels the pixels in the pool to be labeled by considering a different hypothesis about the classification problem. The algorithm then selects the samples showing maximum disagreement between the different classification models in the committee [18]. Query-by-boosting and query-by-bagging are the heuristics used to limit the search. In remote sensing Normalized Entropy Query-by-Bagging (nEQB) and Adaptive Maximum Disagreement (AMD) algorithms have been used for classification between different classes[18].

(ii) *Large margin based heuristics :* The algorithms in this category try to increase the margin between different classes. SVMs are used as the base methods for active learning. The distance to the hyperplane in SVM can be used as the distance measure in the active learning algorithm to decide the value of the decision function. Most popular and researched algorithm in remote sensing is Margin Sampling (MS), where in a multiclass one-against-all setting, the distance to each hyperplane is calculated and the samples are decided on the basis of heuristic calculated as selecting the minimum distance to any of the hyperplanes. The samples with minimum distance are the samples on the margin and are the ones which become the support vectors.

Other algorithms using Multiclass Level Uncertainty (MCLU), Significance Space Construction(SSC) have been discussed in [18], which can be referred for details. It is important to have diversity among the samples used for classification, as it helps to better classify the samples. General diversity based algorithms have also been designed which try to constraint each iteration with diversity among the candidate samples[18].

(iii) *Posterior probability-based heuristics :* In this class of methods, posterior probabilities of class membership are used to rank the candidates. The posterior probability distribution is considered per class for each candidate or change in overall posterior probability. These probabilities are used to select the samples which might maximize the change in posterior probability. Heuristics like KL-Max or Breaking-Ties (BT) can be used [18].

## 6. Classification based on Soft Computing Techniques

6.1 Neural Networks based approaches

Neural networks model the way the training sets would be used to learn. Different types of neural networks are in use now a days. The multiple layers can be used for learning from complex data. Using neural networks for any application requires us to define the architecture of neural network, the activation functions for neurons and the learning algorithm. There are various architectures prominent now a days, for eg, Multi layer perceptron model, self organizing maps, competitive networks, Hopfield. Similarly various activation functions can be defined like sigmoidal, tangent hyperbolic, radial basis etc. The most important aspect of a neural network is the training algorithm that we select, back propagation algorithm, gradient descent has been extensively in NNs. The NNs are quite suitable for remote sensing images, since they have the capability of self-learning, which helps in the NN learning the classification on its own.

Table 1 : Classification based on learning techniques

| Technique | Sub-Category | | Advantages | Disadvantages | Examples |
|---|---|---|---|---|---|
| Unsupervised Change Detection | HCM | | -unsupervised learning<br>- no labeled data is required, less cost | -May be struck in local minima, if initial cluster centroids are such | M.Roy(2013,2014) Wagstaff(2001), L Gomez(2003), Mishra et al(2012) A.Ghosh(2011) |
| | ISODATA | | - no apriori information about data is required<br>- effective at identifying spectral clusters in data<br>- Little user effort required | - time consuming if data is very unstructured<br>-Algorithm can spiral out of control leaving only one class | N. Memarsadeghi (2007) Bo Li.(2010) |
| | FCM | | -able to label overlapping pixels into multiple classes, able to handle uncertainty, imprecision in classification<br>- is distribution free | -Can also be struck in local minima<br>-Extracts clusters of same size, densities and circular shape, fails to perform if the clusters have different shapes<br>-Performance depends on proper selection of fuzzifiers. | M.Roy(2013,2014) Mishra et al(2012) A.Ghosh(2011) |
| | GKC | | -uses adaptive distance norm to measure the distance between clusters<br>-can extract clusters with different shapes<br>- more effective results | - the parameter $\rho_i$ for defining the shape, should be selected properly. | Mishra et al(2012) A.Ghosh(2011) M.Roy(2013,2014) |
| Semi-supervised Change Detection | k-means based | COP-k means | -uses background knowledge in the form of instance level constraints | - a poor decision in the early stage of algorithm, may lead to non assignments of a data in any cluster.<br>- execution time increase as the no. of constraints increase | Wagstaff(2001), M.Roy[2013] |
| | | Seeded k-means | -uses labeled patterns for initialization of cluster centers<br>- rate of convergence is higher | - the original labels are lost after reassignment | M.Roy(2013,2014) |
| | | Constraint k-means | -the labeled patterns are not reassigned to any other cluster, leading to less time requirement | - it may be unable to assign a data point to any of the clusters, as that might violate the constraints | M.Roy(2013,2014) Wagstaff(2001) Bennett et. al (2000) |
| | | Semi-supervised HRMF k-means | - Uses both constraints & an underlying distortion measure for clustering | -can cluster input data only in the form of vectors | Basu(2004), M.Roy(2013,2014) |
| | | Semi-supervised kernel k-means | -discovers clusters with non-linear boundaries, after applying kernel function<br>- can be used for both graph based & vector based input data | -selection of kernel function for a problem domain | Kullis (2005) |
| | Semi supervised SVM (TSVM) | | -unlabeled data helps find a boundary between the classes which is away from dense regions | -to find exact boundary is a NP hard problem<br>- computationally expensive, difficult to work with large data sets | Vapnik (1977) M.Chi(2007) |
| Active Learning Methods | | Committee-based heuristics | -these methods can be applied to any model or combination of models | | Tuia D (2011) |
| | | Large margin based heuristics | -diversity criteria improves the quality of results. | | Tuia D (2011) |
| | | Posterior probability-based heuristics | - If batch size is large, simple heuristics like BT should be used | - Should be avoided if the initial training set is small. | Tuia D (2011) |
| Supervised Change Detection | | | Refer [3], [4], [6] for details | | |

The learning paradigm can be supervised, semi-supervised or un-supervised. In [19] Hopfield type neural networks and self organizing maps have been successfully used for change detection. An overall energy function has been defined for the network and a change detection map is formed as a result. Eliptical Basis NN, MLP have been used in [13] and a multiple classifier system has been designed for classification.

6.2 Fuzzy Logic based approaches

The Change detection problem in remote sensing demands that the pixels be classified as change or no-change class, but there are pixels that overlap between these classes. Fuzzy logic helps us define the membership function for any pixel in different classes, thus it lets us handle the imprecision we have in assigning the pixel to a particular class. In [12], [13] fuzzy clustering has been used for classification using FCM and GKC and fuzzy k nearest neighbour as explained in the previous section.

6.3 Genetic Algorithms (GA) based methods

Genetic algorithms belong to evolutionary computing paradigm. This technique is widely used in optimization problems, where we need to optimize some objective function, which is highly non-linear. Genetic algorithms overcome the problem of local maxima that we face using fuzzy clustering. It helps us to explore a large solution space, to find the optimal solution.

The probable solutions to a problem are coded as chromosomes (representation of solution as strings). Mostly the chromosomes are coded as strings of 0's and 1's. The set of chromosomes is defined as population. A fitness function is associated with the chromosomes(solutions) and a fitness value is evaluated for each particular chromosome. Initially a random set of solutions is generated, and the fitness value for each of these solutions is calculated. Based on this fitness value, a set of two chromosomes is selected using some selection mechanism, known as parent chromosomes. Then two operations *crossover* and *mutation* are applied on these parents(solutions) to generate two offspring's (Children). During *crossover* a crossover point is selected and the strings are exchanged between the parent strings at this point to create the new offspring's with a probability defined as crossover probability. After crossover operation, mutation operation with some defined mutation probability is done on the produced offspring's. During mutation, some of the bits are flipped in the produced offsprings to generate the new offspring's. This operation is repeated on all the parent pairs, thus creating a new population in the process. The above defined steps are repeated till some termination criteria is met.

GA has been used along with fuzzy clustering in [12],[13]. Multiobjective cost function optimization is utilized in [26] for GA and the main advantage is it can also work without computing the difference image and can be used in different types of input satellite images. As per [26] the main drawback was the computational cost, it increase with the size of image.

6.4 Hybrid approaches

As discussed earlier that the techniques NN, FL and GA are complimentary to each other, i.e., we can use a combination of these techniques to create solutions for any problem domain. Mostly NN is combined with FL, where FL is used to represent the data while NN is used for learning and inducting the results.

In change detection a combination of these techniques have been used by many and they have been found to produce better results. In [12] A. Ghosh has combined FL with GA and Simulated Annealing (SA) and analyzed the results thus obtained. Fuzzy clustering has been found to be less costly & simple and perform better when compared to other context sensitive techniques.

# 7. Conclusion

Change detection in remotely sensed images is an important aspect of the applications intended for observing the changes occurring due to natural or man-made processes and restraining the damages being caused by them to Earth's Environment, and planning the utilization of Earths' resources. This paper discussed the basic process of change detection along with the classifications proposed for these methods. The classification based on learning techniques used in softcomputing methods in Change detection has been proposed and discussed.

**Madhu Khurana** has done B.Sc(Comp.Sc) from HansRaj College, Delhi University in 1990, M.Tech(I.T.) from GGSIPU, Delhi in 2009. Currently pursuing Ph.D from JIIT Noida, India. Worked in various companies from 1990 to 1993. Followed by working as a freelance consultant to various National Banks and Trading companies by providing services for system development and maintenance. Academic experience of 11 years, presently working in ABES Engineering College, Ghaziabad for last 7+ years. . Her research interest area is Digital Image Processing and Soft Computing.

**Vikas Saxena** is Associate professor in Dept of CSE&IT at JIIT, Noida. He has more than 12 years of experience in academics. His research interest areas are--DIP, Computer vision and Watermarking. He has more than 40 research paper published in repute international jounals and conferences.